\documentclass[a4paper,conference]{IEEEtran}
\usepackage{cite}

\usepackage{graphicx}
\usepackage{color}
\usepackage{times}
\usepackage{multirow}
\usepackage{subfigure}
\usepackage{epsfig}
\ifCLASSINFOpdf
\else
\fi
\usepackage{amsmath}
\usepackage{amsfonts}
\usepackage{algorithmic}

\usepackage{stfloats}
\begin{document}
%
\title{Concentrated Multi-Grained Multi-Attention Network for Video Based Person Re-Identification}



%
\author{\IEEEauthorblockN{Panwen Hu\IEEEauthorrefmark{1},
Jiazhen Liu\IEEEauthorrefmark{1},
Rui Huang\IEEEauthorrefmark{1} \IEEEauthorrefmark{2}
}
\IEEEauthorblockA{\IEEEauthorrefmark{1} The Chinese University of Hong Kong, Shenzhen}
\IEEEauthorblockA{\IEEEauthorrefmark{2} Shenzhen Institute of Artificial Intelligence and Robotics for Society}
}


\maketitle

\begin{abstract}
Occlusion is still a severe problem in the video-based Re-IDentification (Re-ID) task, which has great impact on the success rate. The attention mechanism has been proved to be helpful in solving the occlusion problem by a large number of existing methods. However, their attention mechanisms still lack the capability to extract sufficient discriminative information into the final representations from the videos. The single attention module scheme employed by existing methods cannot exploit multi-scale spatial cues, and the attention of the single module will be dispersed by multiple salient parts of the person. In this paper, we propose a Concentrated Multi-grained Multi-Attention Network (CMMANet) where two multi-attention modules are designed to extract multi-grained information through processing multi-scale intermediate features. Furthermore, multiple attention submodules in each multi-attention module can automatically discover multiple discriminative regions of the video frames. To achieve this goal, we introduce a diversity loss to diversify the submodules in each multi-attention module, and a concentration loss to integrate their attention responses so that each submodule can strongly focus on a specific meaningful part. The experimental results show that the proposed approach outperforms the state-of-the-art methods by large margins on multiple public datasets.
\end{abstract}

\section{Introduction}
Given an image/video of a person, the goal of the person Re-IDentification(Re-ID) is to retrieve the images/videos of the same person across multiple non-overlapping cameras. In the past few years, various methods have been proposed for the image-based Re-ID task \cite{Loy2013,Zheng2015}. However, the limitation of information contained by a single image usually degenerates the Re-ID performance, especially in dealing with the occlusions. Recently, the image sequence (video) based re-identification has drawn significant attention due to its applications in the intelligent surveillance system. A large number of studies\cite{Zheng2015,Zhou2017,Xu2017,Li2018} proposed different solutions to this task, but it still faces challenges like the variations in camera viewpoints and poses, occlusion, etc.

\begin{figure}[h]
\centering
\subfigure [The attention heatmaps of multi-attention modules trained with (left) and without (right) the proposed concentrated constraint.] {
\label{fig:comparison1}
\begin{minipage}{\columnwidth}
\centering
\includegraphics[width=0.85 \columnwidth, height=8cm]{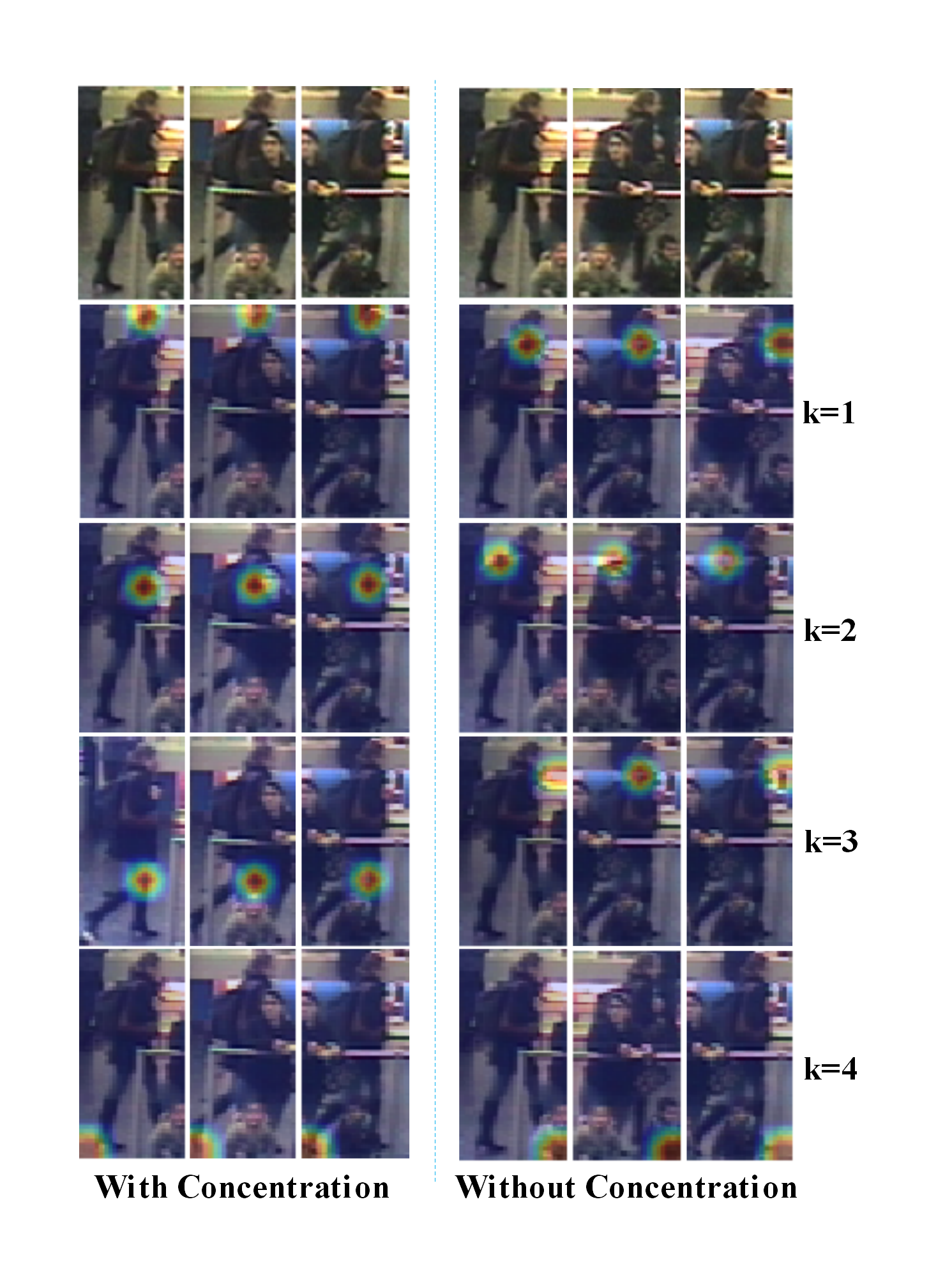}
\vspace{-12 pt}
\end{minipage}
}
\subfigure[The attention heatmaps of different Concentrated Multi-Attention Modules (CMAM), which capture multi-grained cues.] {
\label{fig:multigrained}
\begin{minipage}{\columnwidth}
\centering
\includegraphics[width=0.85 \columnwidth,height=8cm ]{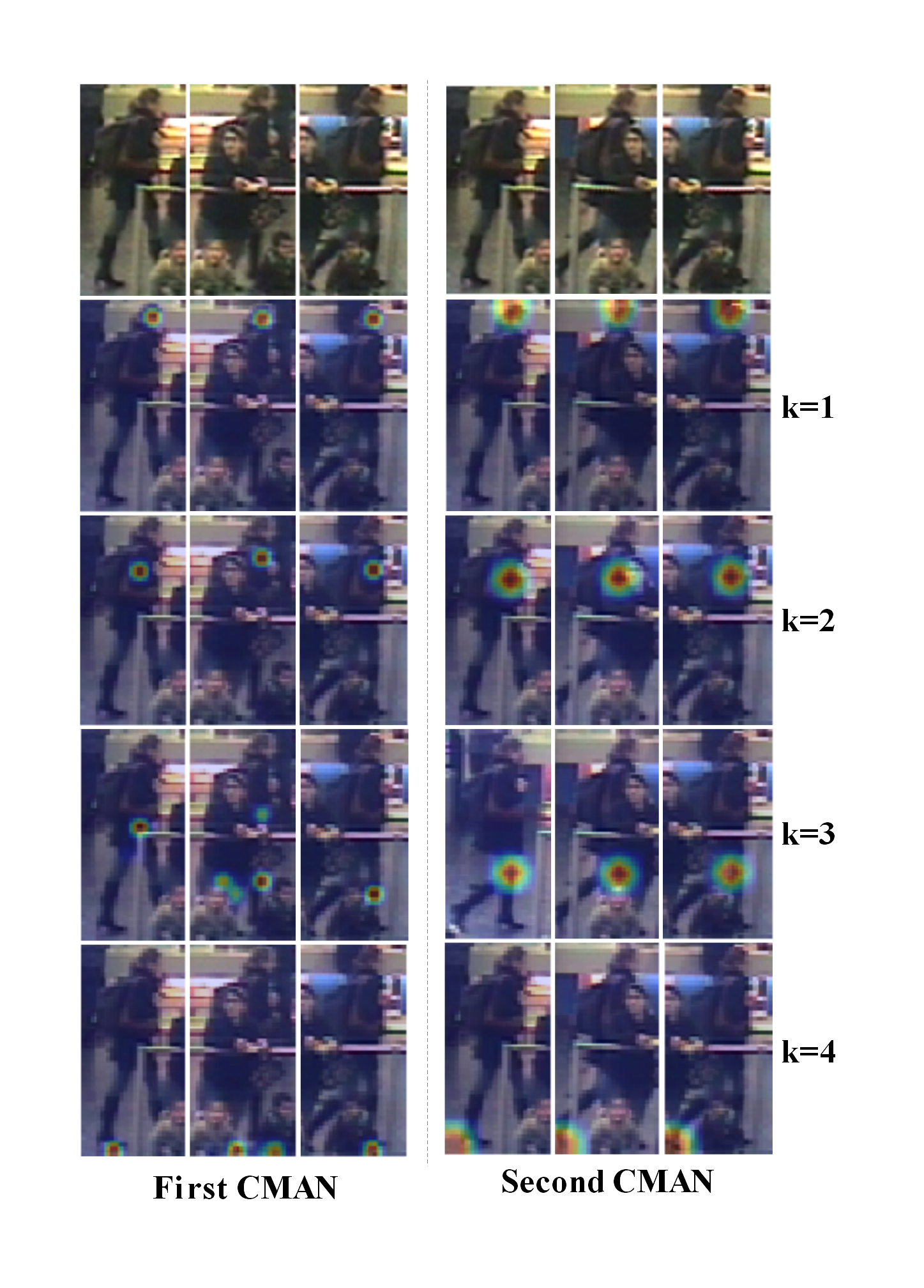}
\vspace{-12 pt}
\end{minipage}
}
\caption{Visualization of attention heatmaps. The first row denotes the original image sequences, and the following four rows visualize heatmaps from different attention submodules. The red color means high response indicating that the corresponding region is more important. }
\label{fig:introducation}
 \vspace{-12 pt}
\end{figure}

A critical step in video-based Re-ID is to learn a mapping function that projects an image sequence into a low-dimensional feature vector, and then we can compare the distances between all feature pairs to achieve re-identification. Thanks to the public large-scale video-based Re-ID datasets \cite{Zheng2016,Wu2018} and the immense capabilities of Neural Networks (NN), the main scheme employed by mainstream works\cite{McLaughlin2016,Li2018} is to train the NN models as the mapping functions in a supervised \cite{Chen2018} or unsupervised manner \cite{Lv2018,Wu2018}, and  the NN based approaches usually achieve better results than the classic methods using hand-crafted features do\cite{Koestinger2012,Liao2015}. Unlike the image-based Re-ID methods where a feature vector represents the content of a single image, the video-based methods are designed to explore the temporal information in an image sequence (e.g. tracklet) besides the spatial contents. Some existing methods \cite{Zhou2017,Xu2017,McLaughlin2016,Yan2016} model the temporal features using the Recurrent Neural Networks (RNN), whereas some other studies\cite{You2016,Liu2017a,Liu2017b} first extracted the feature for each single frame in the image sequence using Convolutional Neural Network(CNN), and then aggregated these features by average pooling to obtain the video representation. However, these methods usually fail while dealing with frequent occlusions or spatial misalignment occurring in the video since they assign equivalent weights to all frames. The final representation of the video is often corrupted by the features of those occluded frames.

Although occlusions can corrupt the video representation, the remaining visible parts of the person can still provide strong cues. To preserve relevant information for Re-ID, recent studies\cite{Zhou2017,Xu2017,Fu2019,Zhang2019,Liu2019a} introduce the attention mechanism to assign different importance weights to different frames or different local parts of a frame to generate a more discriminative representation. However, the single attention module employed by these methods can not express the fine-grained cues of the person like the head, shoulder, feet, etc., except for the coarse visible appearance like the cloth color of the person. Moreover, these methods used the self-attention strategy and did not impose any constraints on the attention modules, which may lead to overfitting easily so that the attention maps cannot reflect the important cues accurately.

To better model the visible cues of the person while there are occlusions, Li et al.\cite{Li2018} proposed the multiple spatial attention model to find the discriminative image regions, and designed a diversity regularization loss to diversify the attentive distributions of multiple attention models. However, the diversity loss only encourages the differences among the attentive distributions, and different attention models may focus on similar regions of the images. For example, as shown in the right part of Fig.\ref{fig:comparison1}, the attention maps of the first three rows (i.e., three different attention models) have high focuses on the upper regions of the images. Since these three attention  models are further diversified by the diversity loss, as a result, some of these models may pay more attention to the meaningless regions for Re-ID.

In this paper, we propose the Concentrated Multi-grained Multi-Attention Network (CMMANet) as shown in Fig.\ref{fig:overall architecture}, which can automatically identify the salient parts of the regions. By imposing a concentration loss (Sec.\ref{subsec:concentrated diversity constrain}) on the Multi-Attention Module(MAM), the attention scores of each attention model will be more concentrated, and different attention models will focus on different regions of the images as shown in the left part of Fig.\ref{fig:comparison1}.

Furthermore, unlike previous methods using single attention module to learn the attention scores from single-scale intermediate features, our proposed architecture has two Concentrated Multi-Attention Module (CMAM, the MAM trained with concentration loss) (Fig.\ref{fig:CMMANet}), which further consists of multiple attention submodules as displayed in Fig.\ref{fig:CMAM}. These two CMAMs allow the network to acquire multi-scale information by exploring the intermediate features with different dimensions. Particularly, the intermediate features fed into first CMAM (the shallower one) have a larger size than those fed into the second CMAM, so the first CMAM perceives the coarse-grained cues from the higher dimensional features, while the second CMAM explores the fine-grained information. As a result, the first CMAM has smaller attentive fields compared to the second CMAM as shown in Fig.\ref{fig:multigrained}. Extensive experiments have been carried out to demonstrate the benefits of combining multi-grained cues.

Moreover, the input videos are usually too long to be completely processed, so we need to sample some frames from each video before feeding the videos into the network. To model different temporal structures of a video, we design a Random Interval Sampling (RIS) strategy (Sec.\ref{subsec:sampling strategy}) to draw the frames. The main contributions of this paper are summarized as follows:
\begin{itemize}
  \item We propose a multi-grain multi-attention architecture for the video-based Re-ID task. Multiple attention submodules in each CMAM identify multiple informative parts of the person, and the multi-grained information collected by two CMAMs enables the final video representation to contain richer information.

  \item In addition to introducing the diversity loss to diversify multiple attention submodules, we propose a concentration loss to integrate the attentive distributions so that each submodule can mainly focus on a specific meaningful part.

   \item We also design a RIS strategy to enrich the training samples. Different frame combinations with varied intervals of each video will be modeled by our method.

\end{itemize}

We conduct various experiments in Sec.\ref{sec:experiment} to demonstrate the effectiveness of the proposed approach on two challenging dataset MARS \cite{Zheng2016} and iLIDS-VID \cite{Wang2014}, and the results show that our method outperform the state-of-the-art methods under multiple common evaluation metrics.

\section{Related Work}
\textbf{Video based person Re-ID.} Video-based Re-ID aims at matching the image sequences of the same person across multiple non-overlapping camera views, and is widely studied recently. Some works \cite{Zhou2017,Xu2017,McLaughlin2016,Yan2016} employed the Recurrent Neural Network(RNN) model to process the sequential images. Yan et al. \cite{Yan2016} used the final states of the RNN as the representations of image sequences. McLaughlin et al.\cite{McLaughlin2016} fed the frame features, which are extracted by the CNN, into the RNN model to incorporate temporal information into each frame, and then applied an average pooling operation to obtain the video representation. Zhou et al.\cite{Zhou2017} used RNN to acquire the attention scores for all frames, and then selected the most discriminative frames from the video. Another popular feature extractor is the convolutional neural network. Liu et al.\cite{Liu2017b} adopted CNN to learn the motion context features from adjacent frames, and the video representation is obtained by applying the average pooling over the features of frames. To avoid the aggregation of frame features, Tran et al.\cite{Tran2015} and Hara et al.\cite{Hara2018} directly explored the video representation using the 3D convolutional network.\\

\noindent \textbf{Attention mechanism in person Re-ID.} The video representations obtained by simply averaging the frame features are usually corrupted by the occlusions in the frames. To handle this problem, the attention mechanism is gaining popularity in the video-based Re-ID community. Liu et al.\cite{Liu2017a} proposed to predict the quality score for each frame of a video using a convolutional subnetwork. In \cite{Xu2017}, Xu et al. proposed a spatial and temporal attention network to select the discriminative regions  from each frame, and the temporal attention scores are obtained by selecting the discriminative frames in the videos. Fu et al.\cite{Fu2019} also proposed a non-parametric attention scheme, where the temporal and spatial importance scores for the pre-divided stripes are computed based on the intermediate feature maps. However, the single attention modules in these methods are trained without specific constraints, and the attention scores usually express mainly the coarse information of the frames. Instead, Li et al.\cite{Li2018} used multiple spatial attention modules aiming at localizing the important parts of the person, and pooled these local fine-grained features over time with temporal attention, while the focus of each attention modules may spread across multiple stripes of a frame. To strengthen the attention on relevant parts of the person, we introduce a concentration constraint to make the focus of each module more compact.

\section{Concentrated Multiple Attention Architecture}
\label{sec:CMA archetecture}

\begin{figure*}[ht]
\centering
\subfigure[The architecture of CMAN] { \label{fig:CMAM}

\begin{minipage}[b]{\textwidth}
\centering
\includegraphics[width=0.9 \columnwidth,height=4cm]{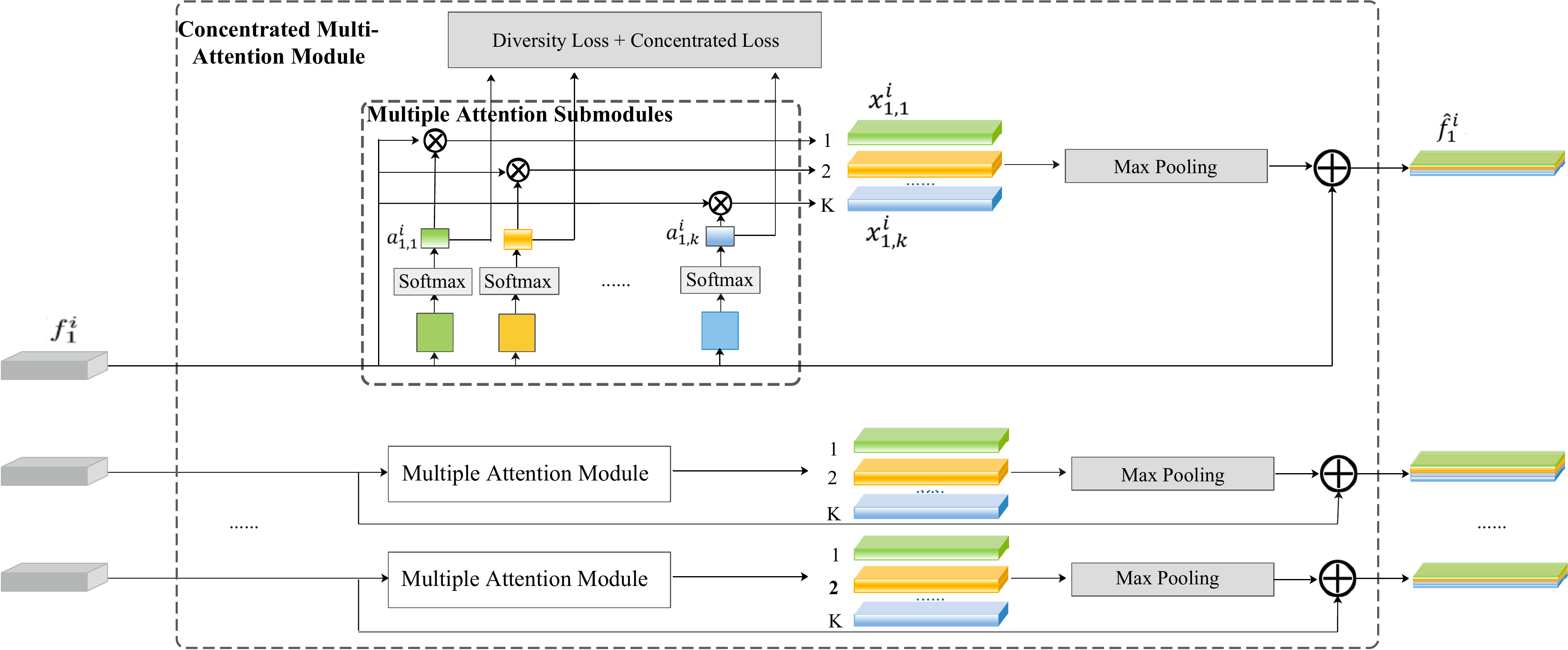}
\end{minipage}
}

\subfigure[The architecture of CMMANet] { \label{fig:CMMANet}

\begin{minipage}[b]{\textwidth}
\centering
\includegraphics[width=0.9 \columnwidth,height=4cm]{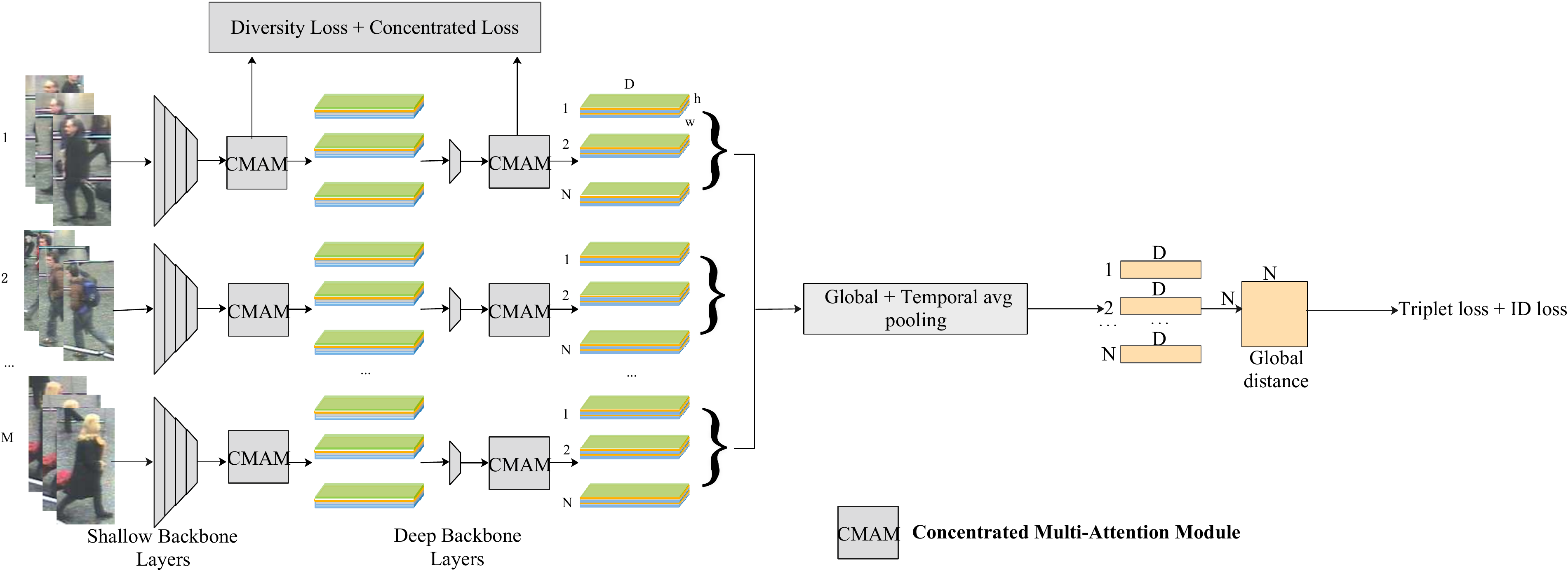}
\end{minipage}
}
\caption{ The framework of the proposed CMMANet. Each CMAM contains multiple attention submodules, which can automatically discover different person parts. Two CMAMs are inserted into different layers of the backbone to exploit multi-grained cues.}
\label{fig:overall architecture}
\end{figure*}

In this section, we will discuss the proposed Concentrated Multi-grained Multi-Attention Network (CMMANet). To enable our network to automatically identify the different discriminative regions of different frames, and strengthen the attention on these regions while extracting features, we use a diversity constraint and the proposed concentration loss (Sec.\ref{subsec:concentrated diversity constrain}) as additional supervised signals during training.
Furthermore, the double CMAMs inserted into different layers of the backbone network ensure that coarse-to-fine cues are aggregated into final video representations. The pipeline of the proposed network architecture is shown in Fig.\ref{fig:overall architecture}, where the CNN backbone network can be replaced with various architecture, like ResNet \cite{He2016}, and Google Inception \cite{Szegedy2016}, etc.

\subsection{Random Interval Sampling strategy}
\label{subsec:sampling strategy}
Previous video-based person Re-ID methods \cite{Ma2017,McLaughlin2016} extracted features from whole videos directly without sampling because their input video sequences are relatively short. With the emergence of large-scale datasets \cite{Wang2014,Zheng2016} where the video sequences are too long to process directly, recent studies \cite{Li2018,Wang2016} proposed the restricted random sampling method which divides the video into fixed number of chunks and then randomly selects a frame from each chunk to constitute the image sequence representing the whole video, however this method only models the long-range temporal structure. To increase the diversity of training samples, we propose a Random Interval Sampling (RIS) strategy, where the ranges that the drawn frames cover on a video are different in different epochs. More precisely, at the beginning of each epoch, the proposed method randomly generates an integer $g$ as the interval between two consecutive frames in the drawn frames. Given the $i$-th input video $V^i = \{I_1^i, I_2^i,\cdots,I_T^i\}$ consisting of $T$ frames, the RIS method generates another integer $s$ ( $\in \left[ 1,T-g*N\right]$ supposing we draw $N$ frames as a training sample) as a sampling start point, so the sample should be $I^i = \{I_{s+g}^i, I_{s+g*2}^i,\dots,I_{s+g*N}^i \}$. For the notation convenience, we use $I^i = \{I_1^i, I_2^i,\cdots,I_N^i\}$ to denote the drawn samples of video $V^i$ for the following sections.

\subsection{Concentrated Multi-Grained Multi-Attention Network}
\label{subsec:CMANet}
The attention mechanism has been widely adopted to tackle the occlusion problem in video-based person Re-ID, whereas computing attention scores from a rigid stripes of the input images lacks global perceptions of the images, and using a single module to acquire attention from a whole image will disperse the attention strength. As a result, the feature representation will easily lose the fine-grained visual cues. To aggregate multi-grained information and make the attention more robust to the occlusions, we propose a multi-grained multi-attention architecture which can automatically identify diverse informative regions of the person from entire images as shown in Fig.\ref{fig:multigrained}, and we further propose a concentrated loss to strengthen the discriminative cues in the extracted features.

As shown in Fig.\ref{fig:overall architecture}, we use the ResNet-50\cite{He2016} as the backbone network. The input image sequence $I^i = \{I_n^i\}_{n=1:N}$ is first passed to the first four layers to extract the per-frame features, which are then fed into the first Multi-Attention Module(MAM) to extract the coarse cues. Note that two MAMs share the same architecture, and the objectives for their attentive distributions are identical (Sec.\ref{subsec:concentrated diversity constrain}). For simplicity, we mainly discuss the calculation in the second MAM, but the same procedure happens in the first MAM as well. The outputs of the first MAM will be used by the last layer of ResNet-50 to generate the intermediate features $\{f^i_1,f^i_2,\cdots,f^i_N\}$,
as a result, each feature has the dimension of $D \times H \times W = 2048 \times 8 \times 4$ with the input image size of $256 \times 128$. Every feature $f^i_n, n\in 1,\cdots,N$ is further fed into the second MAM which consists of $K$ attention submodules to generate $K$ attention weight matrices $\{A_1,A_2,\cdots,A_K\}$ that focus on different regions of the image $I_n^i$. Multiple attention submodules share the same structure, which consists of two convolutional layers $Conv_{1\times1}$ with the kernel size of $1 \times 1$ and a ReLU activation in between, so the response $r_{n,k}$ generated by the $k$-th attention submodule can be written as:
\begin{equation}
\label{eqn:attention response}
  r_{n,k}^i = Conv_{1\times1}(ReLU(Conv_{1\times1}(f_n^i))),
\end{equation}
where $f_n^i \in \mathbb{R}^{d\times h\times w}, r_{n,k}^i \in \mathbb{R}^{h \times w}, n \in 1,\cdots,N, k \in 1,2,\cdots, K$. The inner $Conv_{1\times1}$ reduces the dimension from $D=2048$ to the lower dimensional space $D_1=256$ while keeping the spatial size unchanged, and the outer $Conv_{1\times1}$ further reduces the depth dimension to $1$. To normalize the response intensities to $\left[ 0,1\right]$, we perform a global softmax operation on $r_{n,k}^i$, thus the attentive distribution is
\begin{equation}
\label{eqn:attentive distribution}
  a_{n,k,h,w} = \frac{exp(r_{n,k,h,w})}{\sum_{h=1}^{H} \sum_{w=1}^{W} exp(r_{n,k,h,w})},
\end{equation}
where $a_{n,k} \in \mathbb{R}^{H\times W}$ is the attentive distribution of $k$-th module for image $I_n$.

For each image $I_n$, we generated $K$ attentive distributions which have highest responses on different salient regions of the image, hence the features $X_n^i=\{x_{n,k}^i\}_{k=1:K}$ concerning $K$ salient parts are derived by the Hadamard product between $\{a_{n,k}^i\}_{k=1:K}$ and $f_n^i$ along the depth dimension, i.e.,
\begin{equation}
\label{eqn:weighted features}
  x_{n,k,d}^i = a_{n,k}^i \odot f_{n,d}^i, \quad d \in \{1,\cdots,D\},
\end{equation}
where $d$ denotes the index along the depth dimension. As a result, $x_{n,k}^i$ has the same dimension as $f_{n,k}^i$. To distill the most intensive features from $\{x_{n,k}^i\}_{k=1,K}$ for image $I_n$, we perform an element-wise maximum function to preserve the most informative features. However, the maximum operation will result in the lost of structure information. To mitigate this problem, we introduce a shortcut from the per-frame features $\{f_{n}^i\}_{n=1,N}$, which are then gathered together with $\{x_{n,k}^i\}_{k=1,K}$. Finally, The video representations $F^i$ are obtained by applying a spatial average pooling layer and a temporal average pooling \cite{Fu2019,Liu2019b} on the gathered features$\{\hat{f_{n}^i}\}$. This process is formulated as:
\begin{eqnarray}
  \hat{f_{n}^i} &=& f_{n}^i + \max_k (x_{n,1}^i, \cdots, x_{n,k}^i,\cdots, x_{n,K}^i) \\
  F^i &=& SpatialAvg(TemporalAvg(\hat{f_{1}^i},\cdots,\hat{f_{N}^i}))
\end{eqnarray}
where the $SpatialAvg(\cdot)$ and $TemporalAvg(\cdot)$ are the spatial average pooling and temporal average pooling respectively, the $F^i \in \mathbb{R}^D$ represents the feature vector of video $V^i$. In the Re-ID task, all the representations of gallery videos $\{F^m_g\}_{m=1,M}$ are used to calculate the pair distances with query feature $F^i_q$, and the pair of videos having smallest distance will be treated as the same person.

\subsection{Concentrated Diversity Constraint}
\label{subsec:concentrated diversity constrain}
The attention mechanism is not a new concept in the computer vision community, previous studies\cite{Wang2018,Fu2017,Hu2018} have demonstrated the effectiveness of attention in image processing, whereas these methods did not apply any constraints on the attention modules during training. Directly training multiple attention modules without any constrains would easily produce degenerate features since all the attention modules may be trained to focus on the same salient region. An ideal situation is that different submodules should focus on different receptive regions of the image. In other words, the attentive distributions $\{a_{n,k}^i\}_{k=1,K}$ should be different from each other. Intuitively, to diversify the attentive regions, the network should be trained to maximize the distance between any pair of $a_{n,j}^i$ and $a_{n,k}^i$, which is equivalent to maximizing Eq.\ref{eqn:diversity loss} \cite{Li2018}:
\begin{eqnarray}
\label{eqn:distance}
  D(a_{n,j}^i,a_{n,k}^i) &=& \frac{1}{\sqrt{2}}\|\sqrt{a_{n,j}^i}-\sqrt{a_{n,k}^i}\|_2.
\end{eqnarray}

For computation efficiency, we first flatten each element of $\{a_{n,k}^i\}_{k=1,K}$ to the dimension of $1\times h*w$, and then concatenate them together vertically to form the attention matrix $A^i_n \in \mathbb{R}^{K \times h*w}$. Thus we can formulate the diversity loss $Loss_{div}$ for image $I_n$, which is minimized during training, as follows:
\begin{align}
\label{eqn:diversity loss}
  Loss_{div} &= \sum_{j} \sum_{k,j\neq k} (1 - D^2(a_{n,j}^i,a_{n,k}^i)) \\
  &= \| \sqrt{A^i_n}\sqrt{A^i_n}^T - I \|_F^2
\end{align}
where $I$ is a $K$ dimensional identity matrix. The diversity loss encourages multiple attention submodules to focus on different regions, whereas these submodules may produce similar attentions on the adjacent regions, as a result, some of the submodules will have highest responses on meaningless regions as shown in Fig.\ref{fig:comparison1}, thus the final features can not express the most discriminative information significantly.  To this end, we further propose a concentration loss $Loss_{con}$ to concentrate the attentive intensities of each submodule on a specific region of the entire image. Specifically, we first flatten each $a_{n,k}^i$ and divide it  into $K$ segments $\{\hat{a}_{n,k,l}^i\}_{l=1,K}$ so that each $\hat{a}_{n,k,l}^i \in \mathbb{R}^{h*w/K} $ represent the attention intensities on the $l$-th horizontal stripe of image $I_n$. After processing all the elements of $\{a_{n,k}^i\}_{k=1:K}$, the $K \times K$ concentrated attention matrix $\hat{A}^i_n$ is derived by setting its element $\hat{A}^i_{n,k,l}$ as the summation of $\hat{a}_{n,k,l}^i$, that is:
\begin{equation}
\label{eqn:attentive matrix}
  \hat{A}^i_{n,k,l} = \sum_{j = (l-1)*\Delta}^{l*\Delta} a^i_{n,k,j}
\end{equation}
where the $\Delta$ is the length of $\hat{a}_{n,k,l}^i$, and $a^i_{n,k,j}$ is the j-th element of $a^i_{n,k}$. Intuitively, if every diagonal element of $\hat{A}^i_{n}$ approaches to $1$, the attention of $\hat{a}_{n,k}^i$ will gather on the salient objects in the $k$-th stripe of the original image, thus each submodule focuses on a particular region without losing the knowledge of the entire image, and the representation of discriminative regions can be enhanced in the final video features. To achieve this goal, the network is trained to minimize the concentrated loss $Loss_{con}$, which is written as:
\begin{equation}
\label{eqn:concentrated loss}
  Loss_{con} =  tr(- \log \hat{A}^i_{n} )  \\
\end{equation}
where the $tr(\cdot)$ denotes the trace of the matrix, and $\log$ is the element-wise logarithmic function. By applying $Loss_{con}$, the $k$-th attention module will be more concentrated on the salient parts in $k$-th stripe.

\section{Experiments}
\label{sec:experiment}

\subsection{Datasets And Metrics}
We evaluate our proposed method on two public challenging datasets, MARS \cite{Zheng2016} and iLIDS-VID \cite{Wang2014}. There exist other two pubic benchmarks, PRID-2011 \cite{Hirzer2011} and DukeMTMC \cite{Ristani2016} for video-based Re-ID, but the previous methods \cite{Chen2018,Li2018,Liu2017a} have achieved promising performances on them, while their performances on iLIDS-VID and MARS are still unsatisfied. iLIDS-VID consists of 600 videos of 300 people, and each person has two videos from two cameras respectively. The video length ranges from 23 to 192 frames with an average duration of 73 frames. The challenges mainly result from the occlusions, so it seems more suitable for evaluating our approach. MARS is relatively new and large compared to iLIDS-VID, and consists of 1261 identities and 20715 videos from 6 cameras. Whereas many sequences may have poor quality since the bounding boxes are generated by the DPM detector \cite{Felzenszwalb2009} and the GMMCP tracker \cite{Dehghan2015}, the failures of tracking and detections will affect the Re-ID accuracy. For iLID-VID, we randomly split the probe/gallery identities following the protocol from \cite{Wang2014}. For the MARS dataset, we use the original splits provided by \cite{Zheng2016} which use the predefined 631 people for training and the remaining identities for testing. To quantitatively evaluate our approach, we use the Cumulative Matching Characteristic (CMC) curve and mean Average Precision (mAP) to evaluate the performances as previous studies.

\subsection{Implementation Details}
\label{subsec:implementation}
As mentioned in Sec.\ref{sec:CMA archetecture}, we employ the ResNet-50 pre-trained on ImageNet as the backbone of the CNN network. In the training process,  we randomly select $N=6$ frames for each video using the RIS strategy (Sec.\ref{subsec:sampling strategy}) and then feed them into the network to extract the video features after resizing them to $256 \times 128$. In addition to the diversity and the concentrated objectives which are imposed on the two MAMs, we also introduce the classification loss and the triplet loss\cite{Hermans2017} to constrain the final video representations.  We adopt the Adaptive Moment Estimation (Adam) with the weight decay of 0.0005 to jointly optimize the global branch and temporal branch in an end-to-end manner. The learning rate is initialized to 0.0002, and the batch size is set to 28 due to the limitation of GPU.

\subsection{Comparisons with the State-of-the-arts}
\begin{table}[!htb]
    \begin{minipage}[h]{\columnwidth}
    \small
    \centering
    \caption{Comparison with related methods on iLIDS-VID. * denotes
    those requiring optical flow as inputs. \textcolor[rgb]{1,0,0}{red} indicates the best result, and \textcolor[rgb]{0,0,1}{blue} means the second best result.}
    \begin{tabular}{l | c c c c}
    \hline
    \multirow{2}*{Methods} &\multicolumn{4}{c}{iLIDS-VID} \\
    \cline{2-5}
    &rank-1 &rank-5 &rank-10 &rank-20 \\
    \hline
    LFDA \cite{Pedagadi2013}      &32.9 &68.5 &82.2 &92.6 \\
    KISSME \cite{Koestinger2012}    &36.5 &67.8 &78.8 &87.1 \\
    LADF \cite{Li2013}      &39.0 &76.8 &89.0 &96.8 \\
    STFV3D  \cite{Liu2015}   &44.3 &71.7 &83.7 &91.7 \\
    \hline
    CNN+RNN* \cite{McLaughlin2016}   &58.0 &84.0 &91.0 &96.0 \\
    Seq-Decision \cite{Zhang2018} &60.2 &84.7 &91.7 &95.2 \\
    ASTPN*   \cite{Xu2017}   &62.0 &86.0 &94.0 &98.0 \\
    QAN   \cite{Liu2017a}     &68.0 &86.8 &95.4 &97.4 \\
    RQEN   \cite{Song2018}    &77.1 &93.2 &97.7 &99.4 \\
    STAN \cite{Li2018}    &80.2 &-    &-    &- \\
    Snippet \cite{Chen2018}    &79.8   &91.8 &- &- \\
    Snippet+OF* \cite{Chen2018}   &85.4 &\textcolor[rgb]{0,0,1}{96.7} & \textcolor[rgb]{0,0,1}{98.8} & \textcolor[rgb]{0,0,1}{99.5} \\
    Attribute\cite{Zhao2019}     &83.4 &95.5 &97.7 &\textcolor[rgb]{0,0,1}{99.5}  \\
    SCAN \cite{Zhang2019}     &86.6 & 94.8 & - & 97.1\\

    SCAN+OF* \cite{Zhang2019}  &\textcolor[rgb]{0,0,1}{87.2} & 95.2 & -& 98.1\\
    \hline
    CMMAN(ours) & \textcolor[rgb]{1,0,0}{89.3} & \textcolor[rgb]{1,0,0}{98.7} & \textcolor[rgb]{1,0,0}{100} & \textcolor[rgb]{1,0,0}{100}\\
    \hline
    \end{tabular}
    \label{tbl:ilids}
    \end{minipage}
    \vspace{12 pt}

    \begin{minipage}[h]{\columnwidth}
    \small
    \centering
    \caption{Comparison with related methods on MARS. * denotes those requiring optical flow as inputs. \textcolor[rgb]{1,0,0}{red} indicates the best result, and \textcolor[rgb]{0,0,1}{blue} means the second best result.}
    \begin{center}
    \begin{tabular}{l | c c c c}
    \hline
    \multirow{2}*{Methods} &\multicolumn{4}{c}{MARS} \\
    \cline{2-5}
    &rank-1 &rank-5 &rank-10 &mAP \\
    \hline
    QAN          \cite{Liu2017a}   &73.7 &84.9 &91.6 &51.7 \\
    K-reciprocal  \cite{Zhong2017} &73.9 &-    &-    &68.5 \\
    RQEN        \cite{Song2018}    &77.8 &88.8 &94.3 &71.7 \\
    TriNet      \cite{Hermans2017} &79.8 &91.4 &-    &67.7 \\
    EUG         \cite{Wu2018}     &80.8 &92.1 &96.1 & 67.4 \\
    STAN        \cite{Li2018}     &82.3 &-    &-    &65.8 \\
    Snipped     \cite{Chen2018}   &81.2 &92.1 &-    &69.4 \\
    Snippet+OF*  \cite{Chen2018}   &86.3 &94.7 &\textcolor[rgb]{1,0,0}{98.2} &76.1 \\
    STMP         \cite{Liu2019a}  &84.4 &93.2 &-  & 72.7 \\
    STA          \cite{Fu2019}   &86.3 &\textcolor[rgb]{0,0,1}{95.7} &97.1 &\textcolor[rgb]{0,0,1}{80.8} \\
    SCAN       \cite{Zhang2019}   &86.6 &94.8 & - &76.7 \\
    SCAN+OF*    \cite{Zhang2019}   &\textcolor[rgb]{0,0,1}{87.2} &95.2 & - &77.2 \\
    Attribute    \cite{Zhao2019}   &87.0 &95.4 &- &78.2 \\
    \hline
    CMMAN(ours)    &\textcolor[rgb]{1,0,0}{88.7} & \textcolor[rgb]{1,0,0}{96.2} & \textcolor[rgb]{0,0,1}{97.4} & \textcolor[rgb]{1,0,0}{83.2} \\
    \hline
    \end{tabular}
    \end{center}
    \label{tbl:mars}
    \end{minipage}
    \vspace{ -12 pt}
\end{table}

\textbf{Quantitative Results}
We evaluate our proposed CMMANet on  the iLIDS-VID and MARS datasets, Table.\ref{tbl:ilids} and Table.\ref{tbl:mars} report the performances of our approach on these two datasets. For both datasets, our method attains almost all the highest performances in the metrics of rank-1, rank-5, rank-10, rank-20 and mAP. Specifically, for iLIDS-VID, our approach achieves $89.3\%$ in rank-1 and $100\%$ in rank-10, and improves the state-of-the-art method (SCAN + optical flow) \cite{Zhang2019}  by $2.1\%$ in rank-1, and outperforms the second-best in rank-10 by $1.2\%$. Note that our rank-10 has reached $100\%$, which means the returned first ten most similar gallery videos must contain at least one that belongs to the same person as the query video.  For the MARS dataset, our method attains the best performances in terms of rank-1, rank-5, and mAP. The rank-1 accuracy is improved by $1.6\%$ comparing to SCAN which requires optical flows as extra inputs\cite{Zhang2019}, and the mAP has been raised up by $1.7\%$ in the comparison with the STA\cite{Fu2019}.

\begin{figure}[ht]
\begin{center}
   \includegraphics[width=0.5 \columnwidth, height=7cm]{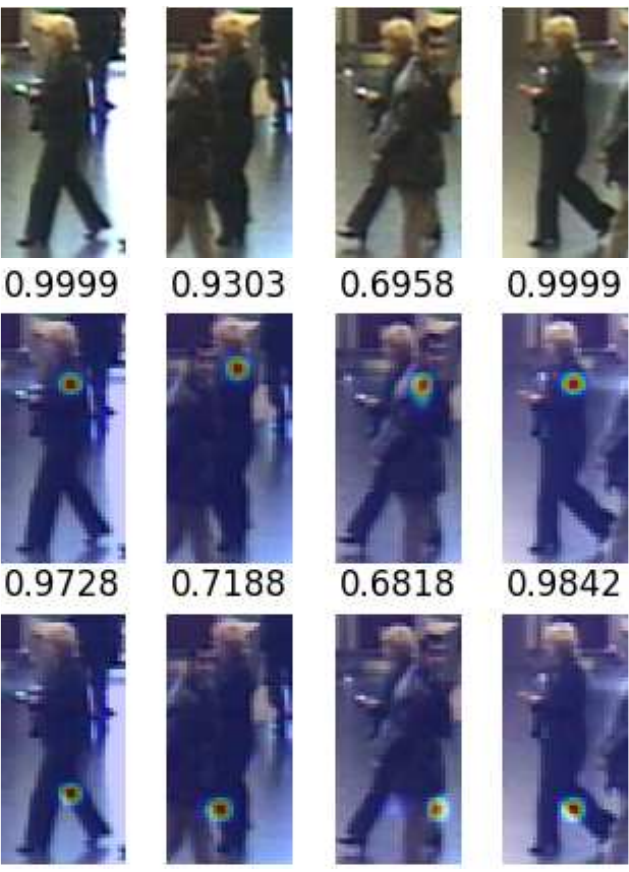}
   \vspace{-12 pt}
\end{center}
   \caption{Visualization of attention heatmaps in different situations. The first row represents the original image sequence. The last two rows show the attention heatmaps from two submodules, and the value at the top of each image means the maximum attention wight in the corresponding heatmap.}
\label{fig:adaptive}
\vspace{-12 pt}
\end{figure}

\noindent \textbf{Visualization of Attention Heatmaps}
To better illustrate the benefits of CMAM in dealing with the occlusions, we visualize the attention heatmaps and the corresponding maximum responses that are collected from two submodules in the first CMAM as shown in Fig.\ref{fig:adaptive}. These two submodules are capable of discovering the parts of the person (our interpretations to these parts are shoulder, leg) even if the occlusions exist. Besides, the attention responses are varied according to different circumstances. For example, when the occlusions appear, such as in the third image of the second row, or the second and third images of the third row, the corresponding submodule will adaptively assign a lower weight to the salient region, and the higher wights will be assigned when the occlusions disappear. The abilities of the attention submodules to automatically discover the salient parts and adaptively weight the parts prevent the video representations from being the corrupted by the occlusions.

\subsection{Ablation Analysis}

\textbf{The influence of sampling strategies} To explore the effectiveness of the RIS strategy, we conducted two experiments on the iLIDS-VID and MARS dataset respectively to make the comparisons. In the first experiment, we implemented the restricted sampling method \cite{Li2018}, and applied it to our CMMANet. We use the RIS approach to preprocess the video in the second experiment while keeping the network and the hyper-parameters unchanged. From the results in Table.\ref{tbl:ablation sampling}, both experiments achieve similar high performance in terms of rank-5, but the RIS method improves the rank-1 by $2.6\%$ and $0.5\%$ on iLIDS-VID and MARS respectively. This confirms that the proposed RIS method can increase the diversity of training samples, and enable the proposed network to generate more discriminative video representations.

\begin{table}[h]\setlength{\tabcolsep}{12pt}
\centering
\footnotesize
\begin{tabular}{c|c|c|c|c}
\hline
\multirow{2}{*}{Sampling Strategies} & \multicolumn{2}{c|}{iLIDS-VID} & \multicolumn{2}{c}{MARS}    \\
\cline{2-5}
 & R1 & R5  &R1 &  R5  \\ \hline
Restricted Sampling \cite{Li2018} & 86.7 & 98.3 & 88.2 & 96.1   \\
Random Interval Sampling          & 89.3 & 98.7 & 88.7 & 96.2 \\ \hline
\end{tabular}
\caption{ The performances of the proposed CMMANet using different sampling strategies.}
\label{tbl:ablation sampling}
\vspace{-12 pt}
\end{table}


\noindent \textbf{The influences of the proposed components} To explore the effectiveness of each component in CMMANet, we conduct several experiments on the iLIDS-VID and MARS dataset with different settings as shown in Table.\ref{tbl:albation ilidsvid} and Table.\ref{tbl:albation mars}. \textbf{Baseline} represents the ResNet-50 backbone network trained with triplet loss $Loss_{trip}$ and ID loss $Loss_{ID}$, \textbf{Baseline + Multi-MAM} use the same network architecture as shown in Fig.\ref{fig:CMMANet}, where two Multi-Attention Modules are placed behind the last two layers of the backbone respectively. To investigate the impact of multi-grained cues, we remove the first MAM from \textbf{Baseline + Multi-MAM} to form \textbf{Baseline + Single MAM} setting, where only single-scale intermediate features are processed by the MAM behind the last layer of the backbone. Furthermore, both \textbf{Baseline + Multi-MAM} and \textbf{Baseline + Single MAM} experiments are conducted with or without the concentration loss \textbf{$Loss_{con}$}  to examine its influences.\\

\begin{table}[h]
\centering
\footnotesize
\begin{tabular}{c|c|c|c}
\hline
\multirow{2}{*}{Methods} & \multicolumn{3}{c}{MARS} \\
\cline{2-4}
 & R1 & R5 & mAP  \\ \hline
\textbf{Baseline}                       & 85.8 & 94.7  &79.4 \\
\textbf{Baseline + Single MAM}     & 87.3 & 95.7   & 82.3  \\
\textbf{Baseline + Single MAM} + $Loss_{con}$   & 88.2 & 96.2   & 82.4  \\
\textbf{Baseline + Multi-MAM}     & 88.0 & 96.0   & 83.0  \\
\textbf{Baseline + Multi-MAM} + $Loss_{con}$      & 88.7 & 96.2  & 83.2  \\ \hline
\end{tabular}
\caption{Component analysis of the proposed CMMANet on MARS. \textbf{Single MAM} means only one MAM is inserted behind the last layer of the backbone, and \textbf{Multi-MAM} means inserting two MAMs behind the last two layers of the backbone. + $Loss_{con}$ denotes the training with concentration loss.}
\label{tbl:albation mars}
\vspace{-12 pt}
\end{table}

\noindent \textbf{Ablation study on MARS} The results of ablation experiments conducted on MARS are shown in Table.\ref{tbl:albation mars}. We can observe that \textbf{Baseline + Single MAM} improves \textbf{Baseline} by $1.5\%$ in rank-1, $1.0\%$ in rank-5, and $2.9\%$ in mAP. This improvement by the MAM indicates its effectiveness and capability to capture useful information. Furthermore,  \textbf{Baseline + Multi-MAM} achieve $0.7\%$ higher performances in both rank-1 and mAP comparing to \textbf{Baseline + Single MAM}, which suggests that the attentions on multi-grained features promote the capability of the network to learn discriminative person representations. We also ran the experiments with and without the proposed $Loss_{con}$. By applying the $Loss_{con}$ during the training, the performances of \textbf{Baseline + Single MAM} and \textbf{Baseline + Multi-MAM} are further raised by $0.9\%$ and $0.7\%$ in rank-1, respectively. It can be concluded that condensing the attentive distribution of each submodule by the $Loss_{con}$ renders the final representations to be more distinguishable.\\

\begin{table}[h]
\centering
\footnotesize
\begin{tabular}{c|c|c|c}
\hline
\multirow{2}{*}{Methods} & \multicolumn{3}{c}{iLIDS-VID} \\
\cline{2-4}
 & R1 & R5 & R10  \\ \hline
\textbf{Baseline}                       & 82.7 & 95.3  &96.7 \\
\textbf{Baseline + Single MAM}     & 84.7 & 97.3   & 99.3  \\
\textbf{Baseline + Single MAM} + $Loss_{con}$   & 85.3 & 98.0   & 100  \\
\textbf{Baseline + Multi-MAM}     & 87.3 & 98.0   & 99.3  \\
\textbf{Baseline + Multi-MAM} + $Loss_{con}$      & 89.3 & 98.7  & 100  \\ \hline
\end{tabular}
\caption{Component analysis of the proposed CMMANet on iLIDS-VID. \textbf{Single MAM} means only one MAM inserted behind the last layer of the backbone, and \textbf{Multi-MAM} means inserting two MAMs behind the last two layers of the backbone. + $Loss_{con}$ denotes the training with concentration loss.}
\label{tbl:albation ilidsvid}
\vspace{-12 pt}
\end{table}

\noindent \textbf{Ablation study on iLDS-VID}
From Table.\ref{tbl:albation ilidsvid}, \textbf{Baseline + Single MAM} improves the \textbf{Baseline} by a large margin $2.0\%$ in rank-1, and \textbf{Baseline + Multi-MAM} further improves the rank-1 result by $2.6\%$ based on \textbf{Baseline + Single MAM}.  Furthermore, we can observe that all experiments with single MAM or double MAMs attain promising results in terms of rank-5 and rank-10. These results reveal that using the proposed MAM can help the network to obtain better features by augmenting the information of the salient regions. Exploring multi-grained cues by multiple MAMs benefits the discrimination of video representations. Moreover, $Loss_{con}$ improves \textbf{Baseline + Single MAM } and \textbf{Baseline + Multi-MAM} by $0.6\%$, $2.0\%$ in rank-1 respectively. We believe that by applying $Loss_{con}$ on the network training, the attention of each submodule in each MAM can be more centralized on a specific meaningful part of the person, and the attention wights for the salient regions are further strengthened. As a result, the video representation will contain stronger discriminative cues.

%

\section{Conclusion}
Developing a mapping function to generate discriminative video representations is a critical step for successful video-base Re-ID. To achieve this goal, the mapping function should learn to avoid the corruption of the video representations by the occlusions, and highlight the discriminative information at the same time. In this work, we propose a new Concentrated Multi-grained Multi-Attention Network (CMMANet) to generate better video representations. Instead of exploring single-scale information, two CMAMs inserted behind the last two layers of the backbone allow the network to exploit coarse-to-fine cues. Furthermore, multiple attention submodules in each CMAM are capable of automatically identifying different discriminative parts of the person in an unsupervised manner. The multi-grained multi-attention design in our approach solves two common problems in video-based Re-ID: determining whether a particular part of the person is occluded or not, and strengthening the visible meaningful parts in the video representation by assigning higher attention weights to them.

To diversify multiple attention submodules in each MAM, and make the attention of each submodule more concentrated on a meaningful body part, we propose a concentration loss, which collaborates with a diversity loss to encourage the submodules to automatically discover a set of non-overlapping salient regions and concentrates the attentive weights of each module on a specific part. Finally, we evaluate our proposed approach on two public datasets and perform a series of experiments to analyze the impact of each component. Our approach outperforms the state-of-the-art methods by a large margin which demonstrates the effectiveness of our network in video-based Re-ID.

%
%
\bibliographystyle{IEEEtran}
\bibliography{egbib}






%

\end{document}